\pdfoutput=1
\documentclass[11pt]{article}

\usepackage[]{acl}
\usepackage{multirow}
\usepackage{amssymb}% http://ctan.org/pkg/amssymb
\usepackage{pifont}% http://ctan.org/pkg/pifont
\newcommand{\cmark}{\ding{51}}%
\newcommand{\xmark}{\ding{55}}%
\usepackage{times}
\usepackage{latexsym}
\usepackage{graphicx}

\usepackage{soul}
\usepackage{color}

\DeclareRobustCommand{\hlcyan}[1]{{\sethlcolor{cyan}\hl{#1}}}
\DeclareRobustCommand{\hlgreen}[1]{{\sethlcolor{green}\hl{#1}}}

\DeclareRobustCommand{\hlyellow}[1]{{\sethlcolor{yellow}\hl{#1}}}
\DeclareRobustCommand{\hlpink}[1]{{\sethlcolor{pink}\hl{#1}}}
\usepackage[T1]{fontenc}
\usepackage[utf8]{inputenc}

\usepackage{microtype}

\title{\textcolor{black}{Zero-shot Code-Mixed Offensive Span Identification through Rationale Extraction}}

\author{Manikandan Ravikiran\textsuperscript{$\dagger$}\thanks{\ \ Corresponding Author}\ , Bharathi Raja Chakravarthi\textsuperscript{$\ddagger$} \\
  \textsuperscript{$\dagger$}Georgia Institute of Technology, Atlanta, Georgia \\
  \textsuperscript{$\ddagger$}Data Science Institute, National University of Ireland Galway \\
  \texttt{mravikiran3@gatech.edu}, \texttt{bharathi.raja@insight-centre.org} \\}

\begin{document}
\maketitle
\begin{abstract}
This paper investigates the effectiveness of sentence-level transformers for zero-shot offensive span identification on a code-mixed Tamil dataset. More specifically, we evaluate rationale extraction methods  of Local Interpretable Model Agnostic Explanations (LIME) \cite{DBLP:conf/kdd/Ribeiro0G16} and Integrated Gradients (IG) \cite{DBLP:conf/icml/SundararajanTY17} for adapting transformer based offensive language classification models for zero-shot offensive span identification. To this end, we find that LIME and IG show baseline $F_{1}$ of 26.35\% and 44.83\%, respectively. Besides, we study the effect of data set size and training process on the overall accuracy of span identification. As a result, we find both LIME and IG to show significant improvement with Masked Data Augmentation and Multilabel Training, with $F_{1}$ of 50.23\% and 47.38\% respectively. \textit{Disclaimer : This paper contains examples that may be considered profane, vulgar, or offensive. The examples do not represent the views of the authors or their employers/graduate schools towards any person(s), group(s), practice(s), or entity/entities. Instead they are used to emphasize only the linguistic research challenges.}

\end{abstract}

\section{Introduction}
Offensive language classification and offensive span identification from code-mixed Tamil-English comments portray the same task at different granularities. In the former case, we classify if the code mixed sentence is offensive or not, while the latter concentrates on extracting the offensive parts of the comments. Accordingly, one could do the former using models of the latter and vice versa. Transformer-based architectures such as BERT \cite{DBLP:conf/naacl/DevlinCLT19}, RoBERTa \cite{DBLP:journals/corr/abs-1907-11692} and XLM-RoBERTa \cite{DBLP:conf/acl/ConneauKGCWGGOZ20} have achieved state-of-the-art results on both these tasks \cite{chakravarthi-etal-2021-findings-shared}. However, often these tasks are treated as independent and model development is often separated.

This paper studies the rationale extraction methods for inferring offensive spans from transformer models trained only on comment-level offensive language classification labels. Such an idea is often vital in the case of code-mixed Tamil-English comments for which span annotations are often costly to obtain, but comment-level labels are readily available. Besides, such an approach will also help in decoding the models' logic behind the prediction of offensiveness.

Accordingly, we evaluate and compare two different methods, namely LIME and IG, for adapting pre-trained transformer models into zero-shot offensive span labelers. Our experiments show that using LIME with pre-trained transformer models struggles to infer correct span level annotations in a zero-shot manner, achieving only 20\% $F_{1}$ on offensive span identification for code-mixed Tamil-English comments. To this end, we find that a combination of masked data augmentation and multilabel training of sentence transformers helps to better focus on individual necessary tokens and achieve a strong baseline on offensive span identification. Besides, IG consistently surpasses LIME even in cases where there are no data augmentation or multilabel training. Overall the contributions of this paper are as follows.

\begin{itemize}
    \item We introduce preliminary experiments on offensive language classification transformer models  for  zero-shot offensive span identification from code-mixed Tamil-English language comments.
    \item We systematically compare LIME and IG methods for zero-shot offensive span identification.
    \item We study the impact of data and training process on offensive span identification by proposing masked data augmentation and multilabel training. 
    \item We further release our code, models, and data to facilitate further research in the field\footnote{The code and data is made available at \url{https://github.com/manikandan-ravikiran/zero-shot-offensive-span}}.
\end{itemize}

The rest of the paper is organized as follows. In Section \ref{methods}, we present LIME and IG methods in brief. Meanwhile section \ref{dataset} and \ref{expsetp} focus on dataset and experimental setup. In section \ref{experiments}, we present detailed experiments and conclude in section \ref{conclusion} with our findings and possible implications on the future work.

\section{Methods}\label{methods}

In this section, we present the two rationale extraction methods LIME and IG used to turn sentence-level transformer models into zero-shot offensive span labelers.

\subsection{Local Interpretable Model Agnostic Explanation (LIME)}
LIME \cite{DBLP:conf/naacl/Ribeiro0G16} is a model agnostic interpretability approach that generates word-level attribution scores using local surrogate models that
are trained on perturbed sentences generated by randomly masking out words in the input sentence. The LIME model has seen considerable traction in the context of rationale extraction for text classification, including work by \citet{thorne-etal-2019-generating}, which suggests that LIME outperforms attention-based approaches to explain NLI models. LIME was also used to probe an LSTM based sentence-pair classifier \cite{lan-xu-2018-neural} by removing tokens from the premise and hypothesis sentences separately. The generated scores are used to perform binary classification of tokens, with the threshold based on $F_{1}$ performance on the development set. The token-level predictions were evaluated against human explanations
of the entailment relation using the e-SNLI dataset \cite{DBLP:journals/corr/abs-1812-01193}.  

Meanwhile, for offensive span identification in English \citet{ding-jurgens-2021-hamiltondinggg} coupled LIME with RoBERTa trained on an expanded training set to find expanded training set could help RoBERTa more accurately learn to recognize toxic span. However, though LIME outperforms other methods, it is significantly slower than Integrated Gradients methods, presented in the next section.

\begin{table*}[!htb]
\centering
\scalebox{0.8}{
\begin{tabular}{|c|c|c|c|c|}
\hline
\textbf{Dataset} & \textbf{\begin{tabular}[c]{@{}c@{}}Number of \\ Train Samples\end{tabular}} & \textbf{\begin{tabular}[c]{@{}c@{}}Number of \\ Test Samples\end{tabular}} & \textbf{\begin{tabular}[c]{@{}c@{}}Offensive Language  \\ Classification\end{tabular}} & \textbf{\begin{tabular}[c]{@{}c@{}}Offensive Span \\ Identification\end{tabular}} \\ \hline
% \textbf{Dravidian CodeMix} & 35139 & 4233 & \cmark & \xmark \\ \hline
\textbf{Offensive Span Identification Dataset} & 4786 & 876 & \xmark & \cmark \\ \hline
\textbf{Mask Augmented Dataset} & 109961 & - & \cmark & \cmark \\ \hline
\textbf{Multilabel Dataset} & 109961 & - & \cmark & \cmark \\ \hline
\end{tabular}}
\caption{Dataset statistics of various datasets used in this work.}
\label{tab:data}
\end{table*}

\begin{table*}[]
\centering
\scalebox{0.8}{
\begin{tabular}{|c|c|}
\hline
\textbf{Steps} & \textbf{Example Output} \\ \hline
\textbf{Offensive Lexicon Creation} & {[}sanghu, thu,thooooo,suthamm,F**k,flop,w*f, p**a,n**y,nakkal{]} \\ \hline
\textbf{Data Sourcing} & \begin{tabular}[c]{@{}c@{}}Last scene vera level love u\\ {[}Last, scene, vera, level, love,u{]}\end{tabular} \\ \hline
\textbf{Mask Generation} & \begin{tabular}[c]{@{}c@{}}{[}0,1, 0, 0, 1,1{]}\\ {[}1,0, 1, 1, 1,1{]}\\ {[}0,0, 0, 0, 1,1{]}\end{tabular} \\ \hline
\textbf{Offensive Word Augmentation} & \begin{tabular}[c]{@{}c@{}}{[}Last, sangh, vera, level, thu, n**y{]}      \\ {[}flop, scene, thooooo, suthamm, F**k, sanghu{]}  \\ {[}Last, scene, p**a, n**y, love, u{]}\end{tabular} \\ \hline
\textbf{\begin{tabular}[c]{@{}c@{}}Position Identification \\ and \\ Multilabel Creation\end{tabular}} & \begin{tabular}[c]{@{}c@{}}Last sangh vera level thu n**y      {[}1,0,1{]}\\ flop scene thooooo suthamm F**k sanghu {[}1,1,1{]}  \\ Last scene p**a n**y love u   {[}0,1,0{]}\end{tabular} \\ \hline
\end{tabular}}
\caption{Various steps in Masked Augmented Dataset and Multilabel Dataset creation with sample outputs.}
\label{tab:maskaug}
\end{table*}

\subsection{Integrated Gradients (IG)}
Integrated Gradients \cite{DBLP:conf/icml/SundararajanTY17} focuses on explaining predictions by integrating the gradient along some trajectory in input space connecting two points. Integrated gradient and its variants are widely used in different fields of deep learning including natural language processing \cite{sikdar-etal-2021-integrated}.  

Specifically, it is an iterative method, which starts with so-called starting baselines, i.e., a starting point that does not contain any information for the model prediction. In our case involving textual data, this is the set exclusively with the start and end tokens. These tokens mark the beginning and the end of a sentence and do not give any information about whether the evaluation is offensive or not. Following this, it takes a certain number of iterations, where the model moves from the starting baseline to the actual input to the model.

This iterative improvement approach is analogous to the sentence creation process wherein each step, we create the sentence word by word and calculate the offensiveness, which in turn gives us the attribution of the input feature. Across its iterations, whenever IG includes an offensive word, we can expect offensive classification prediction to swing more towards offensive class and vice versa. Such behavior will help calculate the attribution of each word in the identified sentence.

\section{Datasets}\label{dataset}

In this section, we present various datasets used in this study. Details on how they are used across different experiments are presented in Table \ref{tab:relations}. Finally, the overall dataset statistics are as shown in Table \ref{tab:data}.

% \subsection{Dravidian Code Mix Dataset}\label{data2}
% This dataset was originally released as part of shared task on offensive language identification \cite{DBLP:journals/corr/abs-2106-09460}. It consists of around 35139 Youtube comments annotated for classification with offensive labels under multiple granularity for training. Out of this, 25425 of are non-offensive while rest (9714) of them are offensive. Additionally, it also has 4233 samples for testing purposes.

\subsection{Offensive Span Identification Dataset}

The Shared task on Offensive Span Identification from Code-Mixed Tamil English Comments \cite{ravikiran-annamalai-2021-dosa} focuses on the extraction of offensive spans from Youtube Comments. The dataset contains 4786 and 876 examples across its train and test set respectively. It consists of annotated offensive spans indicating character offsets of parts of the comments that were offensive.

\subsection{Masked Augmented Dataset}

The data available from both \citet{ravikiran-annamalai-2021-dosa} is minimal, and transformer methods are sensitive to dataset size \cite{xu-etal-2021-optimizing}. Thus we created an additional dataset using Masked Augmentation. Accordingly, the data is generated by using the following steps.

\begin{itemize}
    \item \textbf{Step 1: Offensive Lexicon Creation}: First, we create an offensive lexicon from the train set of offensive span identification datasets. To do this, we do following
    \begin{itemize}
        \item[--] Extract the phrases corresponding to annotated offensive spans from the training dataset of \citet{ravikiran-annamalai-2021-dosa}.
        \item[--] Selecting phrases of size less than 20 characters and word tokenizing them to extract the individual words.
        \item[--] Manually, post-processing these words to ignore words that are not offensive. For example, many phrases include conjunctions and pronouns which are not directly offensive. 
    \end{itemize}
    Accordingly, an offensive lexicon with 2900 tokens is created.
    \item \textbf{Step 2: Data Sourcing}: In this step, we select the dataset used for creating the Masked Augmented dataset. Specifically, we use the 25425 non-offensive comments from Dravidian Code-Mix dataset \cite{DBLP:journals/corr/abs-2106-09460}.
    \item \textbf{Step 3: Mask Generation}: The mask generation is done as follows
    \begin{itemize}
        \item[--] Each of 25425 non-offensive comments was tokenized to create respective \textit{maskable token list}.
        \item[--] Three random binary masks are generated for each of the tokenized non-offensive comments. These binary masks have same length as that of its maskable token list. 
    \end{itemize}
    \item \textbf{Step 4: Offensive Word Augmentation:} Finally, words with a corresponding binary mask of 1 are replaced with words randomly selected from the offensive lexicon from step 1.  Additionally, the spans corresponding to the words that were replaced are saved.
\end{itemize}

Overall, such augmentation resulted in 109961 comments, with 75009 being offensive comments and 34952 non-offensive comments. Table \ref{tab:maskaug} shows an example sentence and masked augmented dataset creation process.

\begin{table*}[]
\centering
\scalebox{0.8}{
\begin{tabular}{|c|cc|}
\hline
\textbf{Experiment  Name} & \multicolumn{1}{c|}{\textbf{Train dataset}} & \textbf{Test dataset} \\ \hline
\textbf{Benchmark} & \multicolumn{2}{c|}{Offensive Span Identification Dataset \cite{ravikiran-annamalai-2021-dosa,toxicspans-acl}} \\ \hline
\textbf{OS-Baseline} & \multicolumn{1}{c|}{Dravidian CodeMix \cite{DBLP:journals/corr/abs-2106-09460}} & \multirow{3}{*}{\begin{tabular}[c]{@{}c@{}}Offensive Span   \\ Identification \\ Dataset\end{tabular}} \\ \cline{1-2}
\textbf{OS-Augmentation} & \multicolumn{1}{c|}{Mask Augmented Dataset} &  \\ \cline{1-2}
\textbf{OS-Multilabel} & \multicolumn{1}{c|}{Multilabel Dataset} &  \\ \hline
\end{tabular}}
\caption{Relationship between the datasets and experiments.}
\label{tab:relations}
\end{table*}

\begin{table}[]
\centering
\scalebox{0.8}{
\begin{tabular}{|c|c|}
\hline
\textbf{Name} & Value \\ \hline
\textbf{$\gamma$} & 0.1 \\ \hline
\textbf{max seq length} & 150 \\ \hline
\textbf{train batch size} & 64 \\ \hline
\textbf{eval batch size} & 64 \\ \hline
\textbf{warmup ratio} & 0.1 \\ \hline
\textbf{learning rate} & 3x$10^{-5}$ \\ \hline
\textbf{weight decay} & 0.1 \\ \hline
\textbf{initializer} & glorot \\ \hline
\end{tabular}}
\caption{Model Hyperparameters for Training Transformers.}
\label{tab:hyper}
\end{table}

\subsection{Multilabel Dataset}
All the previously mentioned datasets are restricted to classification only i.e. they contain a binary label indicating if they are offensive or they have annotated offensive spans. Additionally, these sentences does not explicitly encode any position information of the offensive words, which is useful for training. Work of \citet{Ke2021RethinkingPE} show encoding relative positional information based attention directly in each head often improves the overall result of corresponding down stream task. Similarly \citet{shaw-etal-2018-self} also proposed using relative position encoding instead of absolute position encoding and couple them with key and value projections of Transformers to improve overall results. As such, in this work, to encode position we create a multilabel dataset in which the labels indicate the relative position of offensive words. The multilabel dataset is created as follows.

\begin{itemize}
    \item \textbf{Step 1: Dataset Selection:} We first select the 109961 comments from the Masked Augmented Dataset along with their saved spans.
    \item \textbf{Step 2: Position Identification and Multilabel Creation:} From the identified spans, we check if the offensive spans are present in (a) start of the comment (b) end of the comment and (c) middle of the comment. Depending on presence of offensiveness we create three binary labels. For example in Table \ref{tab:maskaug} for sentence \texttt{Last scene p**a n**y love u} we can see that the offensive word to be present in the center of the sentence. Accordingly we give it a label [0,1,0]. Meanwhile for comment \texttt{Last sangh vera level thu n**y} we can see offensive words to be in center and at the end thus we give label [1,0,1].
\end{itemize}

\section{Experimental Setup} \label{expsetp}

In this section, we present our experimental setup in detail. All of our experiments follow the two steps as explained below. 

\textbf{Transformer Training:} We use three different transformer models, namely Multilingual-BERT, RoBERTa, and XLM-RoBERTa, made available by Hugging Face \cite{DBLP:journals/corr/abs-1910-03771}, as our transformer architecture due to their widespread usage in the context of code-mixed Tamil-English Offensive Language Identification. In line with the works of \citet{DBLP:conf/iclr/MosbachAK21} all the models were fine-tuned for 20 epochs, and the best
performing checkpoint was selected. Each transformer model takes 1 hour to train on a Tesla-V100 GPU with a learning rate of 3x$10^{-5}$. Further, all of our experiments were run five times with different random seeds and the results so reported are an average of five runs. The relationship between the datasets used to train the transformers across various experiments is as shown in Table \ref{tab:relations}. Meanwhile, the model hyperparameters are presented in Table \ref{tab:hyper}.

\textbf{Span Extraction Testing}: After training the transformer models for offensive language identification, we use the test set from the offensive span identification dataset for testing purposes. For LIME, we use individual transformer models' MASK token to mask out individual words and allow LIME to generate 5000 masked samples per sentence. The resulting explanation weights are then used as scores for each word, and tokens below the fixed decision threshold of $\tau=-0.01$ are removed while the spans of the rest of the comments are used for offensive span identification. Meanwhile, for the IG model, for each sentence in the test set, we perform 50 iterations to generate scores for each word and extract the spans in line with LIME.

\section{Experiments, Results and Analysis}\label{experiments}

\begin{table}[!htb]
\centering
\scalebox{0.8}{
\begin{tabular}{|c|c|cc|}
\hline
\multirow{2}{*}{\textbf{Experiments}} & \multirow{2}{*}{\textbf{Model}} & \multicolumn{2}{c|}{\textbf{\begin{tabular}[c]{@{}c@{}}\textbf{$F_{1} (\%)$} \end{tabular}}} \\ \cline{3-4} 
 &  & \multicolumn{1}{c|}{\textbf{LIME}} & \textbf{IG} \\ \hline
\multirow{2}{*}{\textbf{Benchmark}} & \texttt{BENCHMARK 1}  & \multicolumn{2}{c|}{39.834} \\ \cline{2-4} 
 & \texttt{BENCHMARK 2} & \multicolumn{2}{c|}{37.024} \\ \hline
\multirow{3}{*}{\textbf{OS-Baseline}} & BERT & \multicolumn{1}{c|}{26.35} & 44.83 \\ \cline{2-4} 
 & RoBERTa & \multicolumn{1}{c|}{24.26} & 37.01 \\ \cline{2-4} 
 & XLM-RoBERTa & \multicolumn{1}{c|}{22.86} & 43.13 \\ \hline
\multirow{3}{*}{\textbf{OS-Augmentation}} & BERT & \multicolumn{1}{c|}{24.97} & 44.83 \\ \cline{2-4} 
 & RoBERTa & \multicolumn{1}{c|}{26.23} & 44.98 \\ \cline{2-4} 
 & XLM-RoBERTa & \multicolumn{1}{c|}{21.93} & \textbf{50.23} \\ \hline
\multirow{3}{*}{\textbf{OS-Multilabel}} & BERT & \multicolumn{1}{c|}{47.137} & 44.83 \\ \cline{2-4} 
 & RoBERTa & \multicolumn{1}{c|}{\textbf{47.38}} & 35.83 \\ \cline{2-4} 
 & XLM-RoBERTa & \multicolumn{1}{c|}{46.76} & 42.06 \\ \hline
\end{tabular}}
\caption{Consolidated Results on Offensive Span Identification Dataset. All the values represent character level $F_{1}$ measure.}
\label{tab:res_details}
\end{table}

% \begin{table}[!htb]
% \centering
% \scalebox{0.8}{
% \begin{tabular}{|c|c|cc|}
% \hline
% \multirow{2}{*}{\textbf{Experiments}} & \multirow{2}{*}{\textbf{Model}} & \multicolumn{2}{c|}{\textbf{\begin{tabular}[c]{@{}c@{}}\textbf{$F_{1} (\%)$} \end{tabular}}} \\ \cline{3-4} 
%  &  & \multicolumn{1}{c|}{\textbf{LIME}} & \textbf{IG} \\ \hline
% \multirow{2}{*}{\textbf{Benchmark}} & \texttt{BENCHMARK 1}  & \multicolumn{2}{c|}{0.39834} \\ \cline{2-4} 
%  & \texttt{BENCHMARK 2} & \multicolumn{2}{c|}{0.37024} \\ \hline
% \multirow{3}{*}{\textbf{Baseline}} & BERT & \multicolumn{1}{c|}{0.2635} & 0.4483 \\ \cline{2-4} 
%  & RoBERTa & \multicolumn{1}{c|}{0.2426} & 0.3701 \\ \cline{2-4} 
%  & XLM-RoBERTa & \multicolumn{1}{c|}{0.2286} & 0.4313 \\ \hline
% \multirow{3}{*}{\textbf{Data Augmentation}} & BERT & \multicolumn{1}{c|}{0.2497} & 0.4483 \\ \cline{2-4} 
%  & RoBERTa & \multicolumn{1}{c|}{0.2623} & 0.4498 \\ \cline{2-4} 
%  & XLM-RoBERTa & \multicolumn{1}{c|}{0.2193} & \textbf{0.5023} \\ \hline
% \multirow{3}{*}{\textbf{Multilabel}} & BERT & \multicolumn{1}{c|}{0.47137} & 0.4483 \\ \cline{2-4} 
%  & RoBERTa & \multicolumn{1}{c|}{\textbf{0.4738}} & 0.4498 \\ \cline{2-4} 
%  & XLM-RoBERTa & \multicolumn{1}{c|}{0.4676} & 0.5023 \\ \hline
% \end{tabular}}
% \caption{Consolidated Results on Offensive Span Identification Dataset. All the values represent character level $F_{1}$ measure.}
% \label{tab:res_details}
% \end{table}

The consolidated results are presented in Table \ref{tab:res_details}. Each model is trained as an offensive comment classifier and then evaluated for offensive span identification. Though we do not explicitly furnish any signals regarding which words are offensive, we can see an assortment of behaviors across both the rationale extraction methods when trained differently. For reference comparison, we also include two benchmark baseline models from \citet{toxicspans-acl}. \texttt{BENCHMARK 1} is a random baseline model which haphazardly labels 50\% of characters in comments to belong to be offensive inline. \texttt{BENCHMARK 2} is a lexicon-based system, which first extracted all the offensive words from the train samples of offensive span identification dataset \cite{ravikiran-annamalai-2021-dosa}. These words were scoured in comments from the test set during inference, and corresponding spans were noted. We report the character level $F_{1}$ for extracted spans inline with \citet{toxicspans-acl}.

\begin{table*}[]
\centering
\scalebox{0.8}{
\begin{tabular}{|c|c|cc|cc|cc|}
\hline
\multirow{2}{*}{\textbf{Experiments}} & \multirow{2}{*}{\textbf{Model}} & \multicolumn{2}{c|}{\textbf{$F_{1}@30$ (\%)}} & \multicolumn{2}{c|}{\textbf{$F_{1}@50$ (\%)}} & \multicolumn{2}{c|}{\textbf{$F_{1}@>50$ (\%)}} \\ \cline{3-8} 
 &  & \multicolumn{1}{c|}{\textbf{LIME}} & \textbf{IG} & \multicolumn{1}{c|}{\textbf{LIME}} & \textbf{IG} & \multicolumn{1}{c|}{\textbf{LIME}} & \textbf{IG} \\ \hline
\multirow{3}{*}{\textbf{OS-Baseline}} & \textbf{BERT} & \multicolumn{1}{c|}{47.02} & 45.79 & \multicolumn{1}{c|}{32.54} & 50.62 & \multicolumn{1}{c|}{23.27} & 42.48 \\ \cline{2-8} 
 & \textbf{RoBERTa} & \multicolumn{1}{c|}{37.35} & 36.42 & \multicolumn{1}{c|}{32.75} & 42.04 & \multicolumn{1}{c|}{20.56} & 34.95 \\ \cline{2-8} 
 & \textbf{XLM-RoBERTa} & \multicolumn{1}{c|}{43.05} & 51.7 & \multicolumn{1}{c|}{31.54} & 48.69 & \multicolumn{1}{c|}{18.63} & 40.49 \\ \hline
\multirow{3}{*}{\textbf{OS-Augmentation}} & \textbf{BERT} & \multicolumn{1}{c|}{48.45} & 45.79 & \multicolumn{1}{c|}{32.62} & 50.62 & \multicolumn{1}{c|}{21.29} & 42.48 \\ \cline{2-8} 
 & \textbf{RoBERTa} & \multicolumn{1}{c|}{50.21} & 45.86 & \multicolumn{1}{c|}{32.71} & 50.71 & \multicolumn{1}{c|}{22.8} & 42.66 \\ \cline{2-8} 
 & \textbf{XLM-RoBERTa} & \multicolumn{1}{c|}{31.19} & 59.47 & \multicolumn{1}{c|}{27.58} & 57.01 & \multicolumn{1}{c|}{19.17} & 47.19 \\ \hline
\multirow{3}{*}{\textbf{OS-Multilabel}} & \textbf{BERT} & \multicolumn{1}{c|}{45.7} & 45.79 & \multicolumn{1}{c|}{57.15} & 50.62 & \multicolumn{1}{c|}{42.722} & 42.48 \\ \cline{2-8} 
 & \textbf{RoBERTa} & \multicolumn{1}{c|}{58.66} & 45.86 & \multicolumn{1}{c|}{57.19} & 50.71 & \multicolumn{1}{c|}{43} & 42.66 \\ \cline{2-8} 
 & \textbf{XLM-RoBERTa} & \multicolumn{1}{c|}{\textbf{59.84}} & \textbf{59.47} & \multicolumn{1}{c|}{\textbf{57.62}} & \textbf{57.01} & \multicolumn{1}{c|}{\textbf{42.95}} & \textbf{47.19} \\ \hline
\end{tabular}}
\caption{Results across different size of comments.}
\label{tab:scales}
\end{table*}

\begin{table*}[]
\scalebox{0.7}{
\begin{tabular}{|c|c|l|}
\hline
\textbf{Category} & \textbf{Comment Type} & \multicolumn{1}{c|}{\textbf{Examples}} \\ \hline
\multirow{3}{*}{\begin{tabular}[c]{@{}c@{}}Correct \\ Prediction\end{tabular}} & Comments with less than 30 characters & \begin{tabular}[c]{@{}l@{}}\hlgreen{Dei} like poda   anaithu \hlpink{9 p***a}\\  \hlpink{Semma mokka} and as usual a \hlgreen{masala movie}\end{tabular} \\ \cline{2-3} 
 & Comments with 30-50 characters & \begin{tabular}[c]{@{}l@{}} \hlpink{M***u} \hlgreen{adichutu   sagunga} da \hlpink{j***i p***********a}\\  81k views 89k likes YouTube be like \hlpink{W*F}\end{tabular} \\ \cline{2-3} 
 & Comments with greater than 50 characters & \begin{tabular}[c]{@{}l@{}} \hlpink{Old vijayakanth   movie parthathu pola}    irruku..\hlcyan{pidikala}....\\ \hlgreen{Dei} Yappa munjha paarthu \hlpink{Sirichu Sirichu vayiru vazhikuthu} \end{tabular} \\ \hline
\multirow{3}{*}{\begin{tabular}[c]{@{}c@{}}Incorrect \\ Prediction\end{tabular}} & Comments with less than 30 characters & \begin{tabular}[c]{@{}l@{}} \hlgreen{except} last   scene \hlyellow{its a} \hlpink{crap}\\  Movie is \hlyellow{going to be} \hlpink{disaster}\end{tabular} \\ \cline{2-3} 
 & Comments with 30-50 characters & \begin{tabular}[c]{@{}l@{}} Kandasamy and   Mugamoodi\hlyellow{ mixed nu }\hlpink{nenaikre}....\\  \hlyellow{Last la} \hlpink{psycho} ilayaraja nu solitan \end{tabular} \\ \cline{2-3} 
 & Comments with greater than 50 characters & \begin{tabular}[c]{@{}l@{}} All I   \hlgreen{understood} from this video was \hlyellow{Vikram likes} \hlpink{Dosai}..\\ Padam nichiyam oodama poga neriya vaipu \hlgreen{iruku poliye} ! \hlyellow{ Oru} \hlpink{dislike} \hlyellow{ah potu vaipom}\end{tabular} \\ \hline
\end{tabular}}
\caption{Example of correct and incorrect predictions. Blue highlight shows words attributed by LIME. Green highlight shows words attributed by IG. Pink highlight shows words attributed by both LIME and IG. Yellow highlight shows parts of comments annotated in ground truth but not identified by both LIME and IG.}
\label{tab:sample_res}
\end{table*}

\subsection{OS-Baseline Experiments}

Firstly, both benchmarks exhibit high performance, making the task competitive for LIME and IG methods. To start with, we analyze the results of OS-Baseline experiments. From Table \ref{tab:res_details}, we can see that LIME has moderately low performance compared to IG, which either beats the baseline or produces very close results. Analogizing the LIME and IG, we can see that IG has an average difference of 18\% compared to LIME. To understand this, we identify various examples (Table \ref{tab:sample_res}) where LIME fails, and IG performs significantly well and vice versa. Firstly, we can see that LIME explicitly focuses on identifying overtly offensive words only. Besides, we can also see LIME focuses primarily on offensive words, while IG accounts for terms such as \texttt{"Dei"}, \texttt{"understood"}, \texttt{"iruku poliye"} etc.

Accordingly, to comprehend their performance on offensive comments of different sizes, we separate results across (a) comments with less than 30 characters ($F_{1}$@30), (b) comments with 30-50 characters ($F_{1}$@50) (c)  comments with more than 50 characters ($F_{1}$@>50). The results so obtained are as shown in Table \ref{tab:scales}. Accordingly, we find interesting outcomes. Firstly we can see that though LIME has lower $F_{1}$ overall, it tends to show competitive results against IG for comments with less than 30 characters.

With the increase in the comment length, the performance of LIME tends to lower considerably. We believe such behavior of LIME could be because of two reasons (a) surrogate models may not be strong enough to distinguish different classes and (b) dilution of scores due to LIME's random perturbation procedure. With random perturbations, the instances generated may be quite different from training instances drawn from the underlying distribution. Meanwhile, IG is compatible across all the sizes, and in the case of comments with less than 30 and 50 characters, we can see IG to show the result as high as 50\%.

\subsection{OS-Augmentation Experiments}
Since transformers are very sensitive to dataset size, we focus on estimating the impact of dataset size used to train the transformers for offensive comment classification on the performance of LIME and IG, respectively. To this end, we used the Mask Augmented dataset to finetune the transformers and pose the question \textit{Does adding data make any difference?} The various result so obtained are as shown in Table \ref{tab:res_details}. Firstly, for LIME, we see no such drastic difference in $F_{1}$. However, for IG, we can see a significant improvement, especially for RoBERTa and XLM-RoBERTa models. Specifically, we can see the XLM-RoBERTa model to reach an accuracy of 50.23\% with an average of 12\% higher results compared to benchmark models and 7\% compared to OS-Baseline.

Furthermore, analysis of results shows a couple of fascinating characteristics for XLM-RoBERTa. Firstly, we could see many predictions concentrating on words part of the long offensive span annotations. We believe this is because of the ability of the model to learn relations between words in different languages as part of its pretraining, which is not the case with M-BERT and RoBERTa. To verify this again, we separate the results across different comment sizes. From Table \ref{tab:scales} we can see that for longer sized comments, the model tends to outperform M-BERT, RoBERTa when coupled with IG. Meanwhile, LIME has no changes irrespective of used transformers.

\subsection{OS-Multilabel Experiments}
Finally, we analyze the significance of encoding the position of offensive words as part of the training process. To this end, we ask \textit{Does introducing position information as part of the training process improve zero-shot results?}. As such, we use the multilabel dataset to finetune the transformers to obtain results, as shown in Table \ref{tab:res_details}. Firstly, we can see that introducing multiple labels for training has no impact on the overall results of LG. However, we can see that LIME demonstrates a significant gain in overall results. Specifically, with multilabel training, the baseline results improve by ~20\% to 47.38\%.

Furthermore, we can observe an equivalent trend across the different sizes of comments as seen in Table \ref{tab:scales}. In fact, for words of less than 30 and 50 characters, LIME outdoes IG models, which aligns with our hypothesis that the position is helpful. Overall from all the results from Table \ref{tab:res_details}-\ref{tab:scales} we can see XLM-RoBERTa be more suitable for extracting spans, especially with the addition of more data and position information. Meanwhile, IG is consistent in producing explanations irrespective of dataset size or training approach. 

\section{Conclusion }\label{conclusion}

This work examines rationale extraction methods for inferring offensive spans from the transformer model trained for offensive sentence classification. Experiments revealed that approaches such as LIME do not perform as well when applied to transformers directly, attributing to potential issues with surrogate models and perturbation procedures. Meanwhile, we can see IG as the clear front runner for identifying offensive spans in a zero-shot way. We think this is due to the inherent nature of the method, where it focuses on creating the input at the same time learning the reason for offensiveness. 

Besides, we also analyzed LIME and IG under large datasets and incorporated position information in the training process. To this end, we discovered that only augmenting does not improve the performance of LIME. However, when this large data is coupled with labels incorporating position information, both LG and IG improve significantly. Especially LIME prefers this approach with large improvements on $F_{1}$, despite IG outperforming LIME. 

Additionally, we also found XLM-RoBERTa to be a clear winner among the transformer models owing to its intrinsic learning of relationships which potentially helps with comments that are longer size. However, many details were unexplored, including (i) the effect of random perturbations on overall results (ii) the approach to merge attributions of multilabel predictions, which we plan to explore in the immediate future.

\section*{Acknowledgements}
We thank our anonymous reviewers for their valuable feedback. Any opinions, findings, and conclusion or recommendations expressed in this material are those of the authors only and does not reflect the view of their employing organization or graduate schools. The work is the part of the final project in CS7643-Deep Learning class at Georgia Tech (Spring 2022). Bharathi Raja Chakravarthi were supported in part by a research grant from Science Foundation Ireland (SFI) under Grant Number SFI/12/RC/2289$\_$P2 (Insight$\_$2), co-funded by the European Regional Development Fund and Irish Research Council grant IRCLA/2017/129 (CARDAMOM-Comparative Deep Models of Language for Minority and Historical Languages).

\bibliography{anthology,custom}
\bibliographystyle{acl_natbib}

\end{document}